\pdfoutput=1

\documentclass[11pt]{article}

\usepackage{graphicx,booktabs,CJKutf8,newtxmath,multirow,float,tabularx,natbib,makecell}

\usepackage[final]{acl}

\usepackage{times}
\usepackage{latexsym}

\usepackage[T1]{fontenc}

\usepackage[utf8]{inputenc}

\usepackage{microtype}

\usepackage{inconsolata}

\title{Methods for Automatic Matrix Language Determination of Code-Switched Speech}

\author{Olga Iakovenko \and Thomas Hain \\
         The University of Sheffield \\
         \texttt{\{oiakovenko,t.hain\}@sheffield.ac.uk}}

\begin{document}
\begin{CJK*}{UTF8}{gbsn}
\maketitle
\begin{abstract}
Code-switching (CS) is the process of speakers interchanging between two or more languages which in the modern world becomes increasingly common. In order to better describe CS speech the Matrix Language Frame (MLF) theory introduces the concept of a Matrix Language, which is the language that provides the grammatical structure for a CS utterance. In this work the MLF theory was used to develop systems for Matrix Language Identity (MLID) determination. The MLID of English/Mandarin and English/Spanish CS text and speech was compared to acoustic language identity (LID), which is a typical way to identify a language in monolingual utterances. MLID predictors from audio show higher correlation with the textual principles than LID in all cases while also outperforming LID in an MLID recognition task based on F1 macro (60\%) and correlation score (0.38). This novel approach has identified that non-English languages (Mandarin and Spanish) are preferred over the English language as the ML contrary to the monolingual choice of LID.
\end{abstract}

\section{Introduction}
\label{sec:Introduction}

Code-switching (CS) is the process of speakers switching between two or more languages in spoken or written language (Table \ref{tab:cs_example}). Spoken CS data is scarce, and thus models for processing CS speech often yield poor performance in comparison to monolingual models. Given that in many countries CS is widespread (e.g. India, South Africa, Nigeria) \cite{diwan21,ncoko00,rufa83}, it is essential to develop systems for understanding and modelling CS speech. One of the critical tasks in analyzing code-switched speech is determining the matrix language (ML), or the dominant language, which serves as the structural framework for the utterance. Accurate identification of ML is essential for various applications as well as sociolinguistic studies.

\begin{table}[H]
\caption{An example of a CS utterance transcription from the SEAME dataset of colloquial Singaporean language.}
\label{tab:cs_example}
\centering
\begin{tabular}{p{\linewidth}}
{毕业过后  urh  你的  study  life  跟你的  working}\\{life  有什么  difference  吗} \\
\end{tabular}
\end{table}

The linguistic Matrix Language Frame (MLF) theory \cite{myers_scotton97} provides a model for CS production and introduces the concept of a main, i.e. dominant language and a secondary, inserted language in CS utterances. These languages are ML and Embedded Language (EL), respectively. The MLF theory introduces two methods for ML determination:

\begin{enumerate}
    \item \textit{The Morpheme Order Principle} - ML will provide the surface morpheme order for a CS utterance if it consists of singly occurring EL lexemes and any number of ML morphemes
    \item \textit{The System Morpheme Principle} - all system morphemes which have grammatical relations external to their head constituent will come from ML
\end{enumerate}

The morphemes as units within the MLF framework were first introduced by Myers-Scotton in 1997 \cite{myers_scotton97} and were split into \textbf{content} and \textbf{system} morphemes. Some common examples of system morphemes are quantifiers, possessives and tense/aspect determiners, while content morphemes include nouns, pronouns, adjectives, verbs and prepositions.

Matrix language identity (MLID) is the identity of the language providing the grammatical frame for the utterance and it can be defined for both monolingual and CS utterances. Moreover, the existence of ML implies a certain token distribution following the System Morpheme Principle \cite{myers_scotton02} which is highlighted in further Myers-Scotton works in the 4-M model. Overall, MLID provides insight into the grammatical properties of the utterance and a computational implementation of an MLID would be able to reduce the amount of manual annotation.

In this paper three MLID systems for CS text and audio were implemented. MLF theory formulates The Morpheme Order Principle and the System Morpheme Principle, which were implemented into three systems for MLID determination from text (P1.1, P1.2 and P2) and from audio ($MLID_{P1.1}$, $MLID_{P1.2}$ and $MLID_{P2}$). An extensive correlation analysis and comparison of an MLID determination system and a traditional acoustic language identities (LID) were carried out. Recognised MLID and LID from CS texts and audio were compared to ground truth ML annotation and the quality of ML recognition was measured in terms of F1 macro and Matthew's Correlation Coefficient (MCC). To conclude the findings, the distributions of textual LIDs were compared to the textual MLID distributions of the CS data.

The remainder of the paper is structured as follows. The next section reviews the relevant research presented previously on MLF and LID in CS text and speech. The third section provides a detailed description of the methods used. This is followed by a section on experiments, which provides information on datasets, detailed implementation, experiment descriptions as well as a discussion of results. Conclusions summarise and complete the paper.

\section{Related work}
\label{sec:Background}

MLF theory has rarely been used to automatically analyse speech or text. Up until now it was only used for text augmentation \cite{bhat16,yilmaz18,lee19} or for Language Model (LM) adaptation for code-switching. For example additional grammatical information was used during LM construction \cite{adel15,soto19} or a self-supervised training procedure was set up which encouraged generation of CS utterances \cite{gao19}. MLID classification for CS text was carried out in \citet{bullock18pred} where the ML was identified based on the token and system POS majorities.

Simultaneously in the speech processing domain a common technique to separate languages in CS is LID. LID of a whole CS utterance may be performed when CS is regarded as a separate language \cite{metildasagayamaryn20}, in this case the component performs both LID and CS detection. A multilingual ASR system with an utterance-wise LID component as an auxiliary task was tested for CS utterances in \citet{toshniwal18} but the model was not able to generate CS text as a result. LID and language segmentation (LIS) systems make decisions based on similarity to the data they were trained on \cite{muralikrishna21} and, to the best of our knowledge, no study was done to determine if pretrained LID/LIS are able to predict a dominant language in a CS utterance.

Neither MLID nor LID were previously used for CS analysis. Furthermore, ML determination principles were never fully implemented and compared. However, statistical methods were introduced before \cite{guzman17} which can assess the nature of CS. Among the statistical methods only the M-index (Multilingual Index) quantifies the representation of the languages in multilingual corpora. While the M-index is useful to learn about the balance of the token LIDs, it might be insufficient to learn about the utterance LID and MLID distributions.

The above indicates that theoretical methods to identify ML from text exist but previously there was only one attempt to determine MLID which was not based on the two ML determination principles. Furthermore, there are no existing MLID predictors from audio to the best of our knowledge. Therefore the objective of this study is to advance technologies for multilingual understanding and analysis by describing the implementation, comparison and performance of automatic MLID predictors from CS audio and text based on the ML determination principles from the MLF theory.

\section{Principles for ML determination}
\label{sec:Methodology}

\begin{figure*}
\begin{center}
\includegraphics[width=0.7\textwidth]{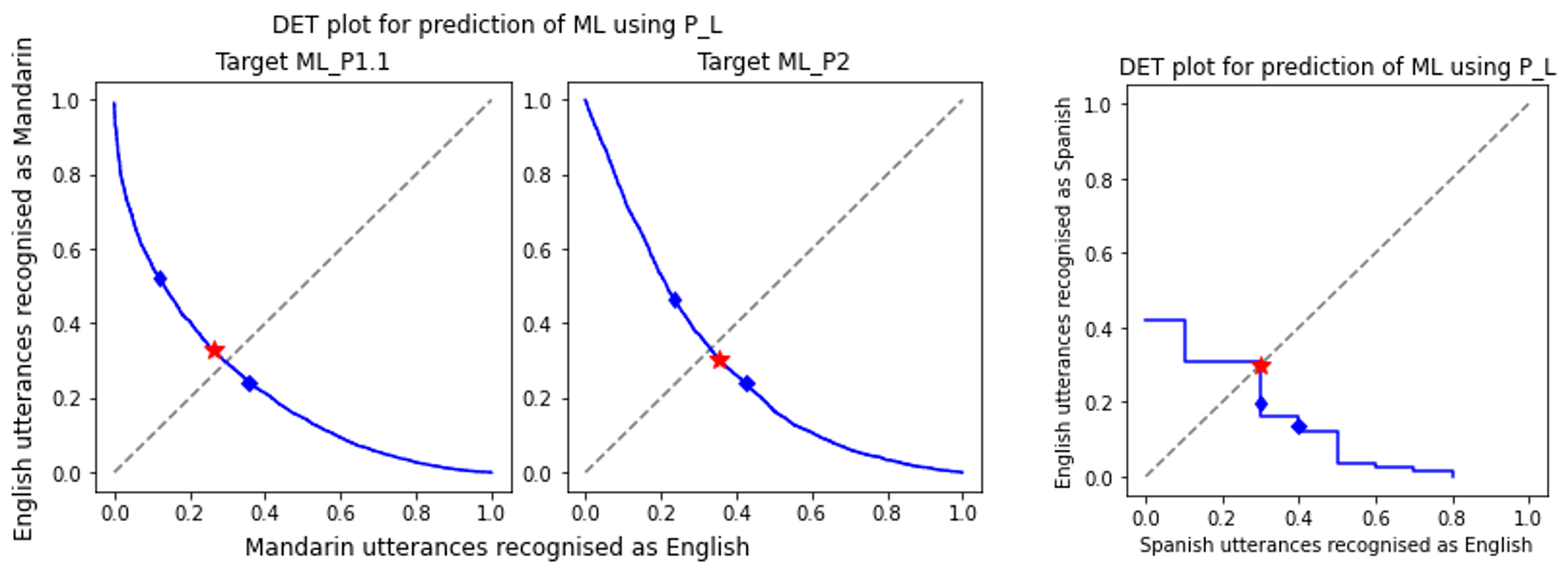}
\caption{Detection error tradeoff (DET) curve for possible $\log\alpha$ values. Thin diamond is the default value of $\log\alpha=0$, thick diamond - result of $\log\alpha$ estimation, red star - ground truth $\log\alpha$.}
\label{fig:det}
\end{center}
\end{figure*}

The MLF theory principles mentioned above have to be implemented in order to compare the main language recognised by an LID system to the MLID. Each of them provides estimates for MLID but is conditioned by different evidence in the utterance. The Morpheme Order Principle is separated into two implementations of the 1st Principle (P1.1 and P1.2). The implementations of the principle deduction as per MLF theory are described below.

\subsection{Principle 1.1: The singleton principle}

The ML provides the context for singly occurring words from the EL, which will be further referred to as "singleton insertions". Although the original principle operates on the level of lexemes, the current implementation operates on the level of words. Suppose there is a CS utterance $\mathbf{y}$ of length $n$, then $(\mathbf{y'}, \mathbf{l'})=((\varepsilon, l_\varepsilon),(y_1,l_1),..,(y_n,l_n),(\varepsilon, l_\varepsilon))$ are morphemes with corresponding language ID labels, $\varepsilon$ is an empty morpheme and $l_\varepsilon$ is an empty language morpheme tag from an empty language $L_\varepsilon$. If $((y_i,l_i),..,(y_j,l_j))$ constitute a word where $0<i<j<n+1$, $\forall k$ that $i<k<j \mid l_k=L_2$ and $l_{i-1},l_{j+1}\in{L_1,L_\varepsilon}$ then the language of the context $L_1$ is the ML while $L_2$ is the embedded language. For example in ``\textbf{哦你} post \textbf{在你的那个} blog`` Mandarin will be ML since it is a context for English singleton insertions.

\subsection{Principle 1.2: The token order principle}

\begin{figure}[H]
\begin{center}
\includegraphics[width=0.5\textwidth]{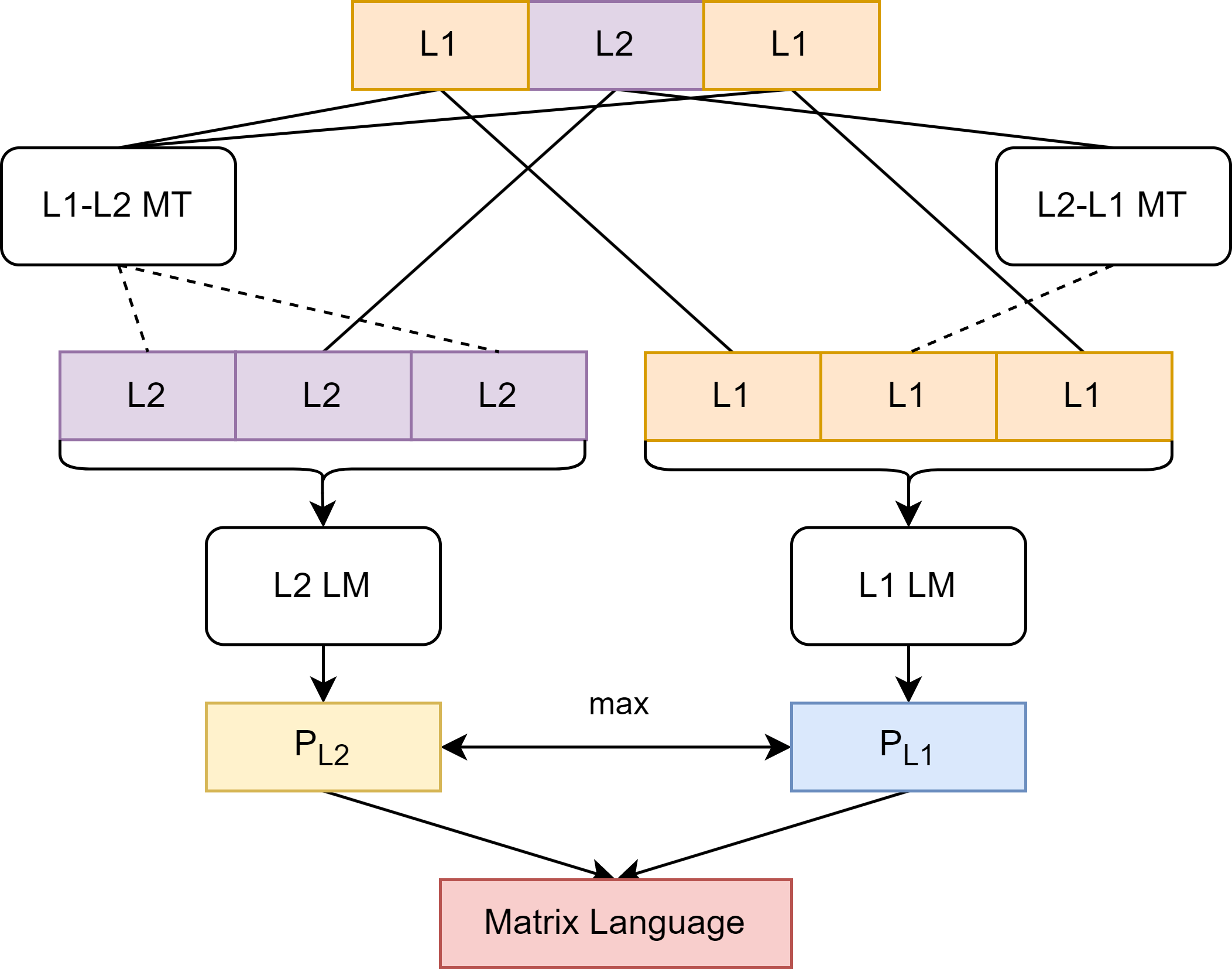}
\caption{Pipeline of the morpheme order-based principle for ML determination P1.2.}
\label{fig:p12}
\end{center}
\end{figure}

\begin{table*}[t]
\caption{Monolingual dataset splits used for LM training in P1.2.}
\label{tab:mono_data}
\centering
\begin{tabular}{c|c|ccc}
{} & {Unit} & \makecell{Callhome\\English} & \makecell{Callhome\\Mandarin} & \makecell{Callhome\\Spanish} \\
\hline
{Token set size} & {tokens} & {6160} & {6853} & {3236} \\
\hline
{Train} & {} & {20029} & {15827} & {19672} \\
{Valid} & {utterances} & {6030} & {3959} & {5500} \\
{Test} & {} & {2609} & {1775} & {2665} \\
\end{tabular}
\end{table*}

In P1.2 the second part of the Morpheme Order Principle is implemented which postulates that the morpheme order is determined by the ML. For example, in ``你觉得我们 speak clear enough \textbf{吗}`` the English translation of the auxiliary Mandarin verb 吗 will never appear at the end of an utterance in English, signifying that Mandarin is ML in this utterance. Assume languages $(L_1, L_2) \subset L$ are present in a bilingual utterance where $L$ are all languages, then the original CS utterance $\mathbf{y}$ can be translated into two monolingual utterances $\mathbf{\hat{y}}_{L_1}$ and $\mathbf{\hat{y}}_{L_2}$. $\mathbf{\hat{y}}_{L_1}$ and $\mathbf{\hat{y}}_{L_2}$ are obtained from the original utterance $\mathbf{y}$ by a Neural Machine Translation (NMT) systems $g_{L_1}$ and $g_{L_2}$. Consider an LM $P(L|\mathbf{y})$ which is used to provide a probability of the utterance belonging to a language $L$, then given the two languages $L_1$ and $L_2$ classification leads to:

\begin{equation}
\frac{P(\mathbf{y}|L_1)P(L_1)}{P(\mathbf{y})} \lesseqgtr \frac{P(\mathbf{y}|L_2)P(L_2)}{P(\mathbf{y})}
\end{equation}

In the above the denominator may be eliminated. The probability $P(\mathbf{y}|L)$ may be estimated using independent monolingual LMs $P_{L}(\mathbf{y})$ and translation $\mathbf{\hat{y}}_{L}$ defined above $P(\mathbf{y}|L) \approx P_{L}(\mathbf{\hat{y}}_{L})$ resulting in following:

\begin{equation}
\frac{P_{L_1}(\mathbf{\hat{y}}_{L_1})}{P_{L_2}(\mathbf{\hat{y}}_{L_2})} \lesseqgtr \alpha
\label{eq:decision_function}
\end{equation}

Where $\alpha$ is the scaling factor for weighing the impact of the models. Taking the log \ref{eq:decision_function} leads to:

\begin{equation}
\log{P_{L_1}(\mathbf{\hat{y}}_{L_1})}-\log{P_{L_2}(\mathbf{\hat{y}}_{L_2})} \lesseqgtr \log{\alpha}
\end{equation}

Assume the difference of language log-probabilities can be expressed in terms of a factor $\alpha$. This factor may be estimated by calculating the expectation of log-probabilities using utterances scored by monolingual LMs:

\begin{equation}
\log{\alpha} = \mathbb{E}\{\log{P_{L_1}(\mathbf{y}_{L_1})}\} - \mathbb{E}\{\log{P_{L_2}(\mathbf{y}_{L_2})}\}
\label{eq:log_alpha}
\end{equation}

All of the above leads to the following decision function:

\begin{equation}
\operatorname{ML} = 
\begin{cases}
L_1, \log{P_{L_1}(\mathbf{\hat{y}}_{L_1})} - \log{P_{L_2}(\mathbf{\hat{y}}_{L_2})} \ge \log{\alpha} \\
L_2, \text{otherwise}
\end{cases}\,
\label{eq:p12_2}
\end{equation}

A visual representation of the resulting algorithm is shown in Figure \ref{fig:p12}.

\subsection{Principle 2: The system word principle}

From the examples in Section \ref{sec:Introduction} it is evident that there exists an overlap of content/system morphemes duality with the traditional content/function words opposition defined in linguistics, although they are not equivalent and the traditional classifications also not strictly distinguishable. Therefore in the implementation of the 2 Principle (P2) for ML determination instead of content/system morpheme duality \cite{myers_scotton02} a content/function Part of Speech (POS) duality is considered.

System POS are identified, namely determiners, auxiliaries, subordinating conjunctions and coordinating conjunctions, while the rest of POS are considered as content POS. ML is determined in a CS utterance if one of the participating languages provided function POS for the utterance and the other language did not. The language that provided the function POS is determined as the ML. Although CS POS taggers exist \cite{feng22, bansal22} none of them are available in open-source and since training a POS tagger is not a goal of this work a monolingual POS tagger is used instead. For example in the utterance ``im okay with \textbf{the} 蛋黄`` determiner ``the`` is used and therefore ML is determined as English.

\section{Experiments}
\label{sec:Experiments}

\subsection{Datasets}

The experiments using the algorithms described above are carried out using monolingual Callhome subsets and 2 CS datasets: SEAME \cite{lyu10} and Miami subcorpus from the Miami-Bangor corpus\footnote{https://biling.talkbank.org/access/Bangor/Miami.html}.

\subsubsection{Monolingual data}

Monolingual LM training for P1.2 was carried out using Callhome datasets for English\footnote{LDC97T14}, Mandarin\footnote{LDC96T16} and Spanish\footnote{LDC96T17}. Pretrained LMs were not used in this work because they do not provide likelihood scoring of morpheme units. The summary of the datasets is presented in Table \ref{tab:mono_data}.

\begin{table*}[t]
\caption{Examples of applying the principles.}
\label{tab:examples}
\centering
\begin{tabular}{c|cccc}
{Utterance} & {baseline} & {ML P1.1} & {ML P1.2} & {ML P2} \\
\hline
{i thought all trains 都是 via jurongeast 去到 pasirris} & {en} & {en} & {en} & {en} \\
{but 他 蛮 zai 的 right} & {en} & {zh} & {en} & {en}  \\
{but 我的 parents 都 没有 sponsor 我} & {zh} & {zh} & {zh} & {en}  \\
{还有 chicken noodles} & {en} & {en} & {en} & {zh}  \\
\end{tabular}
\end{table*}

\subsubsection{CS data}
\label{sec:datasets}

CS spoken language corpora SEAME (120 hours) and Miami (35 hours) are used for analysis and acoustic MLID training. Agreement analysis is carried out for CS utterances from the SEAME and Miami corpora and monolingual SEAME and Miami utterances are used for estimating the scaling factor $\alpha$ (Table \ref{tab:cs_data}). The monolingual subsets of SEAME and Miami are also used for training the mapping from the LID outputs to English, Mandarin and Spanish posteriors ($LID_{map}$). Mandarin characters in the SEAME corpus are word segmented which is helpful when applying P1.1, a principle that operates with words. All the introduced principles require morpheme-level LID tag annotation which is available for Miami and is automatically determined for SEAME based on the script (latin vs logographic). Finally, additional MLID-annotated 91 CS Miami utterances were used to measure the quality of MLID prediction from text and audio. The annotated MLID labels were assigned to the CS utterance transcriptions on the basis of determiner-noun-adjective complexes \cite{couto17}.

\begin{table}[H]
\caption{CS dataset splits.}
\label{tab:cs_data}
\centering
\begin{tabular}{c|c|cc}
{} & {Units} & {SEAME} & {Miami} \\
\hline
{Monolingual} & {} & {53086} & {38401} \\
{CS raw} & {utterances} & {56951} & {2425} \\
{CS annotated} & {} & {-} & {91} \\
\end{tabular}
\end{table}

\subsection{Applying ML Principles to utterance transcriptions}

P1.1 ML only applies to utterance transcriptions with singleton insertions, therefore resulting in a small data coverage: only for 36\% (SEAME) and 60\% (Miami) of the CS data the ML is determined. In P2 POS tags are computed for constituent monolingual islands (segments) of a CS utterance using a pretrained CNN-based POS tagger \cite{svenstrup17}. P2 covered 31\% (SEAME) and 58\% (Miami) of all of the CS examples. Furthermore, a baseline MLID determiner from text is implemented which is based on the token LID count following \citet{bullock18pred}. Examples of running the resulting principles implementations is presented in the Table \ref{tab:examples}.

The implementation of P1.2 includes three components: a Machine Translation (MT) system, a pseudomorpheme tokeniser and a language model (LM). CS utterances are translated word by word using Wiktionary\footnote{https://www.wiktionary.org/} to preserve the token order. The English and Spanish LMs are trained on the tokenised English and Spanish Callhome datasets. The tokenisation was carried out using a stemmer where stem and affix would be separated. For the Mandarin Callhome dataset separate characters are regarded as morphemes. The two Transformer-based \cite{vaswani17} LMs with 2 layers, 2 attention heads per layer are trained for 25 epochs with negative log-likelihood loss on one 3080 Nvidia GPU for 1 hour. Validation and test perplexities for the three languages are presented in Table \ref{tab:ppl}.

\begin{table}[H]
\caption{Perplexities calculated for the validation and test subsets of monolingual Callhome data.}
\label{tab:ppl}
\centering
\begin{tabular}{c|ccc}
{} & {English} & {Mandarin} & {Spanish} \\
\hline
{Valid} & {48.97} & {94.98} & {57.76} \\
{Test} & {57.61} & {98.16} & {52.30} \\
\end{tabular}
\end{table}

Moreover, a preliminary experiment is carried out to evaluate if the trained LMs have the ability to detect the original word order (WO) among its permuted variants (up to 20 word permutations). The sequence of tokens for which the probability was the highest was chosen as the predicted original WO. Comparing the sequence with chosen WO to the original WO leads to 37\% accuracy for SEAME and 60\% for Miami.

\begin{table*}[t]
  \caption{Experimental results for SEAME. First three columns and last three rows (P1.1, P1.2 and P2) refer to the ML determination principles from text. "Coverage" row presents the percentage of all CS examples being processed. "\% English" row displays the percentage of utterances recognised as "English" LID or MLID. MCC Baseline refers to the word LID majority implementation \cite{bullock18pred}. "$LID$" is a pretrained LID system, "$LID_{map}$" column is a mapping trained on monolingual utterances from SEAME. $MLID_{P1.1}$, $MLID_{P1.2}$ and $MLID_{P2}$ are trained mappings similar to $LID_{map}$ but trained on CS data and labels generated from transcriptions by corresponding principles. $MLID_{P1.1}$, $MLID_{P1.2}$ and $MLID_{P2}$ contain correlation values with the target MLID determined from text (\textit{italic}) and correlations with other MLID targets.}
  \label{tab:seame}
  \centering
  \begin{tabular}{c|ccc|cc|ccc}
    \toprule
    {} & P1.1 & P1.2 & P2 & $LID$ & $LID_{map}$ & $MLID_{P1.1}$ & $MLID_{P1.2}$ & $MLID_{P2}$ \\
    \midrule
    Coverage & 36\% & 100\% & 31\% & 100\% & 100\% & 100\% & 100\% & 100\% \\
    \% English & 24\% & 46\% & 49\% & 18\% & 43\% & 42\% & 44\% & 45\% \\
    \midrule
    MCC Baseline & 0.99 & 0.28 & 0.69 & 0.33 & 0.33 & \textbf{0.5}& 0.38& 0.46\\
    MCC P1.1 & 1 & 0.36 & 0.83 & 0.41 & 0.5 & \textit{0.67} & 0.47 & \textbf{0.52} \\
    MCC P1.2 & 0.36 & 1 & 0.31 & 0.09 & 0.14 & \textbf{0.17} & \textit{0.3} & 0.16 \\
    MCC P2 & 0.83 & 0.31 & 1 & 0.33 & 0.45 & \textbf{0.49} & 0.4 & \textit{0.6} \\
    \bottomrule
  \end{tabular}
\end{table*}

\begin{table}[H]
\caption{Outcomes of $\alpha$ estimation. "-$\alpha$ MCC" is the correlation measured between the MLID determined by the unscaled P1.2 approach and MLID labels from other principles (+ true MLID labels for Miami). "+$\alpha$ MCC" are the correlation measurements with the scaled P1.2.}
\label{tab:alpha}
\centering
\begin{tabular}{c|cc|ccc}
{} & \multicolumn{2}{c|}{SEAME} & \multicolumn{3}{c}{Miami} \\
{} & {P1.1} & {P2} & {P1.1} & {P2} & {true} \\
\hline
-$\alpha$ MCC & 0.31 & \textbf{0.33} & 0.36 & 0.08 & \textbf{0.41} \\
+$\alpha$ MCC & \textbf{0.36} & 0.31 & \textbf{0.38} & \textbf{0.09} & 0.37 \\
\end{tabular}
\end{table}

Outputs of the pre-trained monolingual LMs have different probability distributions, therefore, as described in Section \ref{sec:Methodology}, the factor $\alpha$ is used to allow for scale changes. $\alpha$ is derived from expectations of the probabilities yielded on monolingual examples and their translations following Equation \ref{eq:log_alpha}. As a result of $\alpha$ estimation the MCC of SEAME P1.1/P1.2, Miami P1.1/P1.2 and Miami P2/P1.2 has increased (Table \ref{tab:alpha}). Additionally, a "true" $\alpha$ value is calculated using ground truth MLID for Miami and P1.1 and P2 MLID for SEAME, and they are compared to the estimated $\alpha$. DET plots and highlighted thresholds in Figure \ref{fig:det} demonstrate that by using the estimated $\alpha$ the amount of False Positives (FP) and False Negatives (FN) becomes more balanced for SEAME. For Miami the $\alpha$ estimation does not lead to more balanced FP and FN but this improvement is not observed due to the limited test set and other reasons which will be discussed later (Section \ref{sec:correlation}).

\subsection{Language Identification}

If one assumes that is a "dominant" language that most acoustically resembles the spoken CS utterance, then a conventional LID system can be used as an ML determiner. An ECAPA-TDNN \cite{desplanques20} model pretrained on Voxlingua-107 \cite{valk20} was used to automatically detect the dominant language from audio data (Table \ref{tab:seame} and \ref{tab:miami}, column $LID$). The ECAPA-TDNN model was trained to recognise a large number of languages. In order to limit the models to binary task a mapping function was trained from the outputs based on a fully-connected neural network (Multi-Layer Perceptron, MLP) classifier. The mapping function is trained to map 107 language output posteriors to the binary output of the languages participating in CS. LID is a challenging task for accented data such as monolingual subsets from SEAME and Miami but still achieves 82\% and 79\% F1-macro respectively on cross-validation among 5 splits.

\subsection{ML identification from audio}

One can train an MLP mapping model using the LID posterior distribution to also predict P1.1, P1.2 and P2 from audio. Due to the different coverage rates of P1.1, P1.2 and P2 of the CS data the amount of training data would vary greatly: 16582 for P1.1, 43068 for P1.2 and 23868 for P2. The resulting systems will be further referred to as $MLID_{P1.1}$, $MLID_{P1.2}$ and $MLID_{P2}$.

\subsection{Correlation analysis}

The agreement between the implemented principles is measured using the MCC metric since the MLID generated by the principles are not human annotation and are automatically generated. F1-macro is computed only in cases when the human-annotated Miami subset is compared to the MLID approaches.

\begin{table*}[t]
  \caption{Experimental results for Miami. "$LID_{map}$" column is a mapping trained on monolingual utterances from Miami. "F1-macro true" and "MCC true" are the metric values when comparing the outputs of the systems to ground truth ML annotation for Miami.}
  \label{tab:miami}
  \centering
  \begin{tabular}{c|ccc|cc|ccc}
    \toprule
    {} & P1.1 & P1.2 & P2 & $LID$ & $LID_{map}$ & $MLID_{P1.1}$ & $MLID_{P1.2}$ & $MLID_{P2}$ \\
    \midrule
    Coverage & 60\% & 100\% & 58\% & 100\% & 100\% & 100\% & 100\% & 100\% \\
    \% English & 45\% & 31\% & 31\% & 43\% & 30\% & 31\% & 42\% & 31\% \\
    \midrule
    F1-macro true & 100\% & 67\% & 93\% & 56\% & 53\% & 56\% & \textbf{60\%} & 56\% \\
    MCC true      & 1.0   & 0.37 & 0.86 & 0.27 & 0.35 & 0.27 & 0.24          & \textbf{0.38} \\
    \midrule
    MCC Baseline & 0.99& 0.28& 0.67& 0.59& 0.81& \textbf{0.83}& 0.42& 0.8\\
    MCC P1.1      & 1     & 0.38 & 0.81 & 0.45 & 0.42 & \textit{0.85} & 0.43          & \textbf{0.82} \\
    MCC P1.2      & 0.38  & 1    & 0.09 & 0.26 & \textbf{0.34} & 0.35 & \textit{0.53}          & 0.34 \\
    MCC P2        & 0.81  & 0.09 & 1    & 0.7  & 0.86 & \textbf{0.87} & 0.51          & \textit{0.82} \\
    \bottomrule
  \end{tabular}
\end{table*}

\subsubsection{Correlation between P1.1, P1.2 and P2}

\begin{table}[H]
  \caption{Correlation values for SEAME with unlabelled sentences given a third "unknown MLID P1.1", "unknown MLID P1.2" and "unknown MLID P2" class labels for the three principle implementations accordingly. This approach ensures 100\% coverage of all utterances sacrificing the MCC.}
  \label{tab:seame_with_unk}
  \centering
  \begin{tabular}{c|ccc}
    \toprule
    {} & P1.1 & P1.2 & P2 \\
    \midrule
    MCC P1.1 & 1 & 0.03 & 0.09 \\
    MCC P1.2 & 0.03 & 1 & 0.1 \\
    MCC P2 & 0.09 & 0.1 & 1 \\
    \bottomrule
  \end{tabular}
\end{table}

P1.1, P1.2 and P2 were applied to CS text data and the agreement analysis is presented in Table \ref{tab:seame} and Table \ref{tab:miami} for SEAME and Miami respectively in the first three columns. P1.1 and P2 have to meet certain conditions to be applied, therefore they do not have full coverage of CS data: 36\% and 31\% for SEAME, 60\% and 58\% for Miami. Measuring the correlations only for the utterances for which the MLID is determined is performed to measure the agreement of the implementations of the linguistic principles. Higher correlation indicates that the principles in agreement may be used to generate ground truth MLID labels for downstream tasks. Calculating MCC with unknown labels leads to extremely low correlation due to a large portion of unknown labels which makes it impossible to assess the slight changes in correlation of the labeled data (Tables \ref{tab:seame_with_unk} and \ref{tab:miami_with_unk}).

Among the three principles P1.1 and P2 have the greatest correlation (0.82 for SEAME and 0.81 for Miami), P1.1/P1.2 demonstrates less correlation (0.36 and 0.38), while the least correlation is observed between P1.2 and P2 (0.31 and 0.09). P1.1 and the baseline have almost identical behavior which is expected (0.99 and 1.0), whereas less correlation is observed of the baseline with P2 (0.69 and 0.67) and P1.2 (0.28 and 0.28).

\begin{table}[H]
  \caption{Correlation values for Miami with unlabelled sentences given the "unknown ML" class.}
  \label{tab:miami_with_unk}
  \centering
  \begin{tabular}{c|ccc}
    \toprule
    {} & P1.1 & P1.2 & P2 \\
    \midrule
    F1-macro true & 55\% & 67\% & 48\% \\
    MCC true      & 0.44   & 0.37 & 0.56 \\
    \midrule
    MCC P1.1      & 1     & 0.1 & 0.25 \\
    MCC P1.2      & 0.1  & 1    & 0.16 \\
    MCC P2        & 0.25  & 0.16 & 1 \\
    \bottomrule
  \end{tabular}
\end{table}

\begin{table*}[t]
\caption{Distributions of languages in CS corpora. Utterance level LID for monolingual subsets is in the "Utterance LID" row, token level LID for CS is in the "Token LID" row and utterance level textual ML for CS are in rows P1.1/P1.2/P2.}
\label{tab:cs_distributions}
\centering
\begin{tabular}{c|cc|cc}
{} & \multicolumn{2}{c|}{SEAME} & \multicolumn{2}{c}{Miami} \\
{} & {English} & {Mandarin} & {English} & {Spanish} \\
\hline
{Utterance LID (mono)} & {54\%} & {46\%} & {68\%} & {32\%} \\
{Token LID (CS)} & {42\%} & {58\%} & {66\%} & {34\%} \\
\hline
{P1.1 (CS)} & {23\%} & {77\%} & {45\%} & {55\%} \\
{P1.2 (CS)} & {44\%} & {56\%} & {31\%} & {69\%} \\
{P2 (CS)} & {49\%} & {51\%} & {31\%} & {69\%} \\
\end{tabular}
\end{table*}

The high correlation values for P1.1 and P2 prove that the MLF framework can reliably predict the structure and behavior of CS text. This enables to use the MLIDs generated by the rule-based principles as pseudo-labels in applications.

\subsubsection{Correlation of P1.1/P1.2/P2 and the acoustic LID/MLID}
\label{sec:correlation}


The ML determined from CS text is compared to the LID computed from the corresponding audio. The procedure for the LID experiments is described in the previous subsection. Columns 4 and 5 in Tables \ref{tab:seame} and \ref{tab:miami} show the amount of correlation between MLID derived from text and the recognised LID classes. The same columns for Miami in Table \ref{tab:miami} also include F1 macro and MCC for an annotated MLID subset. Training $LID_{map}$ on the monolingual utterances seems to increase the MCC (from 0.27 to 0.35) but decrease the F1 macro (56\% from 53\%) for the CS Miami data.

Suppose a conventional LID system determines the dominant language in a CS audio based on the majority of time the language is spoken. Then the true annotation may be approximated by counting the textual token LIDs in a CS utterance (Baseline). However, correlation analysis shows that $MLID$ systems are better predictors of the token LID majority (columns 4-5 vs 6-8 row MCC Baseline in Tables \ref{tab:seame} and \ref{tab:miami}).

Further experimentation comprises of comparing $MLID_{P1.1}$, $MLID_{P1.2}$ and $MLID_{P2}$ with P1.1, P1.2 and P2. Upon observing the results for SEAME data $MLID_{P1.2}$ leads to overall highest value out-of-domain MCC scores (0.47+0.4=0.87) for textual principles P1.1 and P2. A similar inspection of the Miami results shows the biggest MCC scores for $MLID_{P1.1}$ (0.35+0.87). For the annotated subset of Miami data $MLID_{P1.2}$ leads to the biggest F1 macro among all systems (60\%), while $MLID_{P2}$ leads to the biggest MCC score (0.38).

\begin{figure}[H]
\begin{center}
\includegraphics[width=0.4\textwidth]{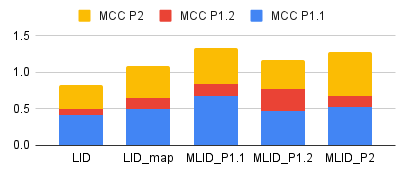}
\caption{Correlations between acoustic $LID$ and $MLID$ outputs and textual P1.1, P1.2 and P2 for CS SEAME data. Each bar segment represents the amount of correlation for a LID or MLID model with textual principles, therefore the whole bar represents the sum of the correlations.}
\label{fig:seame_bar}
\end{center}
\end{figure}

\begin{figure}[H]
\begin{center}
\includegraphics[width=0.4\textwidth]{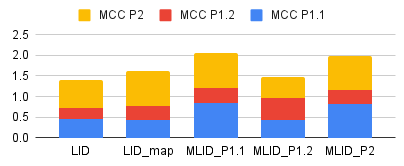}
\caption{Correlations between acoustic $LID$ and $MLID$ outputs and textual P1.1, P1.2 and P2 for CS Miami data.}
\label{fig:miami_bar}
\end{center}
\end{figure}

Lastly, the MCC scores between the textual and acoustic MLID determiners are summed up for every $LID$/$MLID$ approach (Figures \ref{fig:seame_bar} and \ref{fig:miami_bar}). Results show that the correlation of the proposed approaches with the MLID is higher than in $LID$ systems in all occasions apart from $MLID_{P1.2}$ for CS Miami data. The latter is due to the word order of the English and Spanish languages being similar in contrast to English/Mandarin CS. This leads to the morpheme order having a better discrimination power in the case of English/Mandarin CS than in English/Spanish CS.

\subsubsection{P1.1/P1.2/P2 distribution analysis}

At the last step of analysis the distributions of languages are measured on utterance level LID for monolingual (Utterance LID row in Table \ref{tab:cs_distributions}), token level LID for CS (Token LID row in Table \ref{tab:cs_distributions}) and utterance level textual MLID for CS (P1.1/P1.2/P2 rows in Table \ref{tab:cs_distributions}). The numbers reveal that although the majority of the monolingual utterances are English in both corpora (54\% for SEAME and 68\% for Miami), it is not the preferred ML when CS occurs in the utterance for all principles. The token LID distribution also does not seem to be correlated with the choice of the ML in these corpora. In SEAME there seems to be a strong preference towards using Mandarin as an ML (77\%) when EL insertions are single words (P1.1). The preference is not as strong for Spanish in the CS Miami subset (55\%) but it is still a big difference in comparison to the monolingual distributions (32\%). P1.2 and P2 show a similar distribution of MLIDs with the numbers indicating the preference of speakers to use the non-English language as the grammatical frame for a CS utterance.

\section{Conclusion}
\label{sssec:Conclusion}

To the best of our knowledge this is the first work that precisely carries out the Matrix Language (ML) determination of a code-switched (CS) utterance based on the Matrix Language Frame (MLF) theory and that compares Matrix Language Identity (MLID) to acoustic Language Identity (LID). Three methods for ML determination in text and audio are implemented using the ideas and the concepts of the MLF theory \cite{myers_scotton97}. An extensive correlation analysis of the MLID systems from text and speech is carried out. A pretrained LID system $LID$ is adapted to the data by training a mapping function $LID_{map}$, while also mapping functions $MLID_{P1.1}$, $MLID_{P1.2}$, $MLID_{P2}$ for MLID are trained. $MLID$ consistently outperforms $LID$ for ML determination from audio based on Matthew's Correlation Coefficient (MCC). Comparing the results to the ground truth ML annotation shows that the trained $MLID_{P1.2}$ and $MLID_{P2}$ outperform $LID$ in terms of F1-macro and MCC respectively.  Finally, this approach reveals that despite English dominating as the utterance LID for the monolingual utterances, non-English (Mandarin or Spanish) languages set the grammatical frame for CS utterances.

The proposed approaches can be used for accurate automatic analysis of CS text and audio. It can provide insight into the nature of CS for whole datasets but also separate speakers and even utterances. Further work will explore the usefulness of the MLID implementations in Natural Language Processing and Automatic Speech Recognition (ASR) applications, namely in language and dialogue modelling and also in end-to-end multitask ASR the MLID component will be used as a part of the ASR setup. Additionally, further development of P2 is required where the system morphemes would be automatically determined from a given set of CS data rather than using a closed set of POS tags.

\section*{Limitations}
The main limitation of the method is related to data availability: there is limited ML-annotated CS data openly available to date. Therefore it is problematic to assess the quality of ML classification. ML identity can be determined in CS data using the P1.1 but the principle can only be applied in case of singleton EL insertions. Since there is no ML annotation, correlation was measured for most of the experiments which is difficult to assess. Finally, although providing valuable insight into the CS data, the usefulness of the method is yet to be tested in NLP and ASR applications.

\section*{Acknowledgements}
Many thanks to Maria del Carmen Parafita Couto and Diana Carter for sharing the ML annotation for the Miami-Bangor corpus. We would also like to thank Tom Pickard and Carolina Scarton for the very helpful comments on the paper. This work was supported by Engineering and Physical Sciences Research Council [grant number 2676033].

\bibliography{refs}

\end{CJK*}
\end{document}